\documentclass[10pt,twocolumn,letterpaper]{article}

\usepackage{iccv}
\usepackage{times}
\usepackage{epsfig}
\usepackage{graphicx}
\usepackage{amsmath}
\usepackage{amssymb}
\usepackage{color}
\usepackage{subcaption}
\usepackage{multirow}
\usepackage{booktabs}
\usepackage{multirow}

\usepackage[pagebackref=true,breaklinks=true,letterpaper=true,colorlinks,bookmarks=false]{hyperref}

\newcommand{\cmd}[1]{{\tt #1}}

\iccvfinalcopy 

\def\iccvPaperID{XXXX} 

\ificcvfinal\fi
\begin{document}

\title{Learning Data Augmentation Strategies for Object Detection}
\author{Barret Zoph\thanks{Equal contribution.}, Ekin D. Cubuk\footnotemark[1], Golnaz Ghiasi, Tsung-Yi Lin, Jonathon Shlens, Quoc V. Le \\
Google Research, Brain Team \\
\texttt{\{barretzoph, cubuk, golnazg, tsungyi, shlens, qvl\}@google.com}
}

\maketitle

\begin{abstract}
Data augmentation is a critical component of training deep learning models. Although data augmentation has been shown to significantly improve
image classification, its potential has not been thoroughly investigated for 
object detection.
Given the additional cost for annotating images for object detection, data augmentation may be of even greater importance for this computer vision task.
In this work, we study the impact of data augmentation on object detection. We first demonstrate that data augmentation operations borrowed from image classification may be helpful for training detection models, but the improvement is limited.
Thus, we investigate how learned, specialized data augmentation policies
improve generalization performance for detection models.
Importantly, these augmentation policies only affect training and leave a trained model unchanged during evaluation.
Experiments on the COCO dataset indicate that an optimized data augmentation policy improves detection accuracy by more than +2.3 mAP, and allow a single inference model to achieve a state-of-the-art accuracy of 50.7 mAP. Importantly, the best policy found on COCO may be transferred unchanged to other detection datasets and models to improve predictive accuracy. For example, the best  augmentation policy identified with COCO improves a strong baseline on PASCAL-VOC by +2.7 mAP. Our results also reveal that a learned augmentation policy is superior to state-of-the-art architecture regularization methods for object detection, even when considering strong baselines. Code for training with the learned policy is available online.
\footnote{\texttt{\href{}{github.com/tensorflow/tpu/tree/master/models/ \\ official/detection}}}

\end{abstract}

\section{Introduction}
Deep neural networks are powerful machine learning systems that work best when trained on vast amounts of data.  To increase the amount of training data for neural networks, much work was devoted to creating better data augmentation strategies~\cite{baird1992document,simard2003best,krizhevsky2012imagenet}. In the image domain, common augmentations include translating the image by a few pixels, or flipping the image horizontally. Most modern image classifiers are paired with hand-crafted data augmentation strategies \cite{krizhevsky2012imagenet,szegedy2015going,he2016deep,hu2017squeeze,zoph2017learning}. 

\begin{figure}[t]
\centering
\begin{minipage}[c]{0.35\linewidth}
\includegraphics[width=0.9\linewidth]{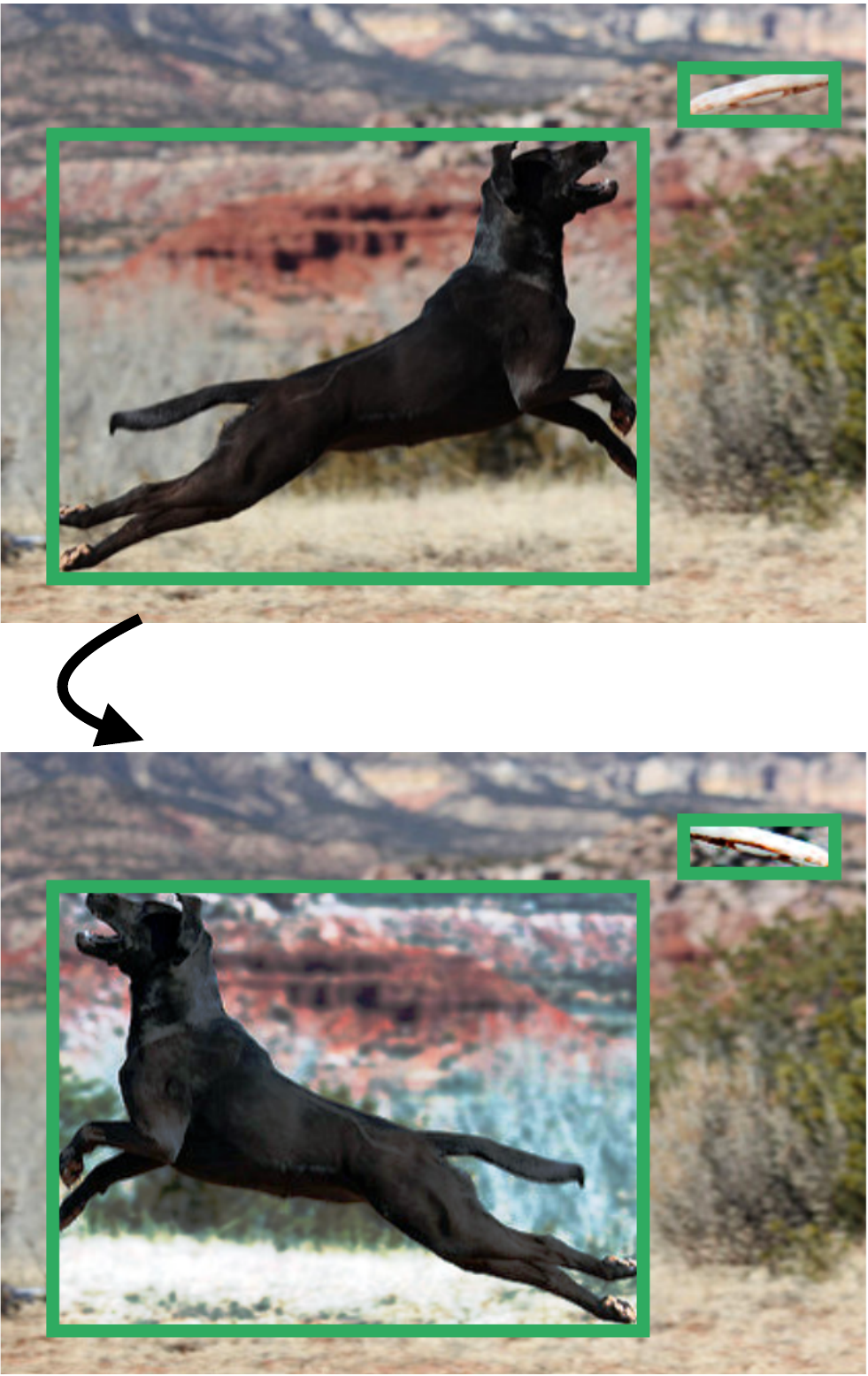}
\end{minipage}%
\begin{minipage}[c]{0.65\linewidth}
\vspace{12pt}
\includegraphics[width=1.0\linewidth]{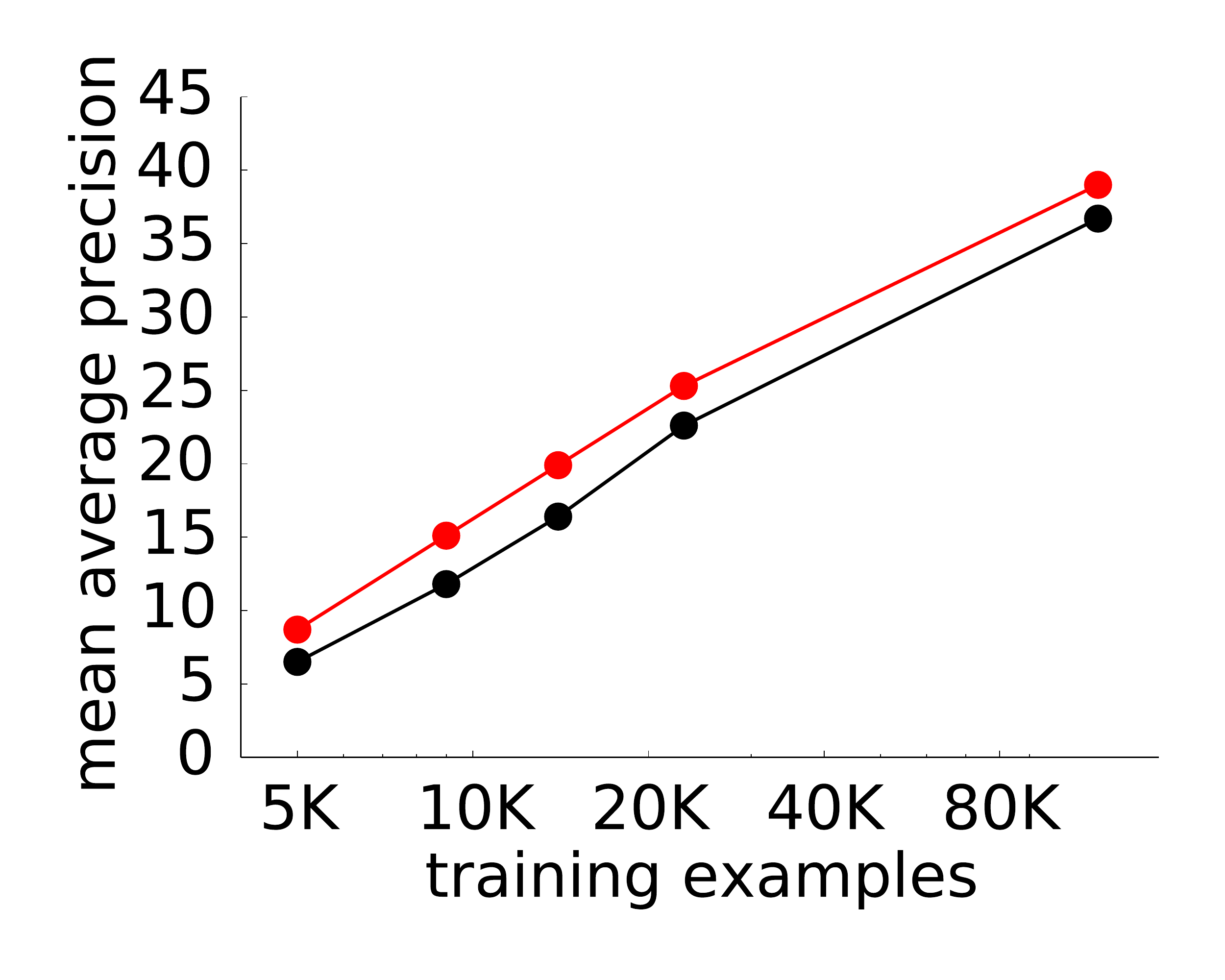}
\end{minipage}%
\caption{\textbf{Learned augmentation policy systematically improves object detection performance}. Left: Learned augmentation policy applied to example from COCO dataset \cite{lin2014microsoft}. Right: Mean average precision for RetinaNet \cite{lin2017focal} with a ResNet-50 backbone on COCO \cite{lin2014microsoft} with and without learned augmentation policy (red and black, respectively). }

\label{fig:marketing}
\end{figure}
 
Recent work has shown that instead of manually designing data augmentation strategies, learning an optimal policy from data can lead to significant improvements in generalization performance of image classification models~\cite{lemley2017smart, tran2017bayesian, devries2017dataset,perez2017effectiveness,mun2017generative,zhu2017data,antoniou2017data,sixt2016rendergan, ratner2017learning,cubuk2018autoaugment}. For image classification models, data can be augmented either by learning a generator that can create data from scratch~\cite{perez2017effectiveness,mun2017generative,zhu2017data,antoniou2017data,sixt2016rendergan}, or by learning a set of transformations as applied to already existing training set samples~\cite{cubuk2018autoaugment,ratner2017learning}. For object detection models, the need for data augmentation is more crucial as collecting labeled data for detection is more costly and common detection datasets have many fewer examples than image classification datasets. It is, however, unclear how to augment the data: Should we directly reuse data augmentation strategies from image classification? What should we do with the bounding boxes and the contents of the bounding boxes? 

In this work, we create a set of simple transformations that may be applied to object detection datasets and then transfer these transformations to other detection datasets and architectures. These transformations are only used during training and not test time. Our transformations include those that can be applied to the whole image without affecting the bounding box locations (e.g. color transformations borrowed from image classification models), transformations that affect the whole image while changing the bounding box locations (e.g., translating or shearing of the whole image), and transformations that are only applied to objects within the bounding boxes. As the number of transformations becomes large, it becomes non-trivial to manually combine them effectively. We therefore search for policies specifically designed for object detection datasets. Experiments show that this method achieves very good performance across different datasets, dataset sizes, backbone architectures and detection algorithms. Additionally, we investigate how the performance of a data augmentation policy depends on the number of operations included in the search space
and how the effective of the augmentation technique varies as dataset size changes. 

In summary, our main contributions are as follows:
\begin{itemize}
\item Design and implement a search method to combine and optimize data augmentation policies for object detection problems by combining novel operations specific to bounding box annotations.
\item Demonstrate consistent gains in cross-validated accuracy across a range of detection architectures and datasets.  In particular, we exceed state-of-the-art results on COCO for a single model and achieve competitive results on the PASCAL VOC object detection.
\item Highlight how the learned data augmentation strategies are particularly advantageous for small datasets by providing a strong regularization to avoid over-fitting on small objects.
\end{itemize}

\section{Related Work}
Data augmentation strategies for vision models are often specific dataset or even machine learning architectures. For example, state-of-the-art models trained on MNIST use elastic distortions which effect scale, translation, and rotation~\cite{simard2003best,ciregan2012multi,wan2013regularization,sato2015apac}. Random cropping and image mirroring are commonly used in classification models trained on natural images~\cite{WRN2016,krizhevsky2012imagenet}. 
Among the data augmentation strategies for object detection, image mirror and multi-scale training are the most widely used~\cite{girshick2018detectron}. Object-centric cropping is a popular augmentation approach \cite{liu2016ssd}. Instead of cropping to focus on parts of the image, some methods randomly erase or add noise to patches of images for improved accuracy~\cite{cutout2017,zhong2017random,ghiasi2018dropblock}, robustness~\cite{yin2019afourier,ford2019adversarial}, or both~\cite{lopes2019improving}. In the same vein, ~\cite{wang2017fast} learns an occlusion pattern for each object to create adversarial examples. In addition to cropping and erasing, \cite{dwibedi2017cut} adds new objects on training images by cut-and-paste.

To avoid the data-specific nature of data augmentation, recent work has focused on learning data augmentation strategies directly from data itself. For example, Smart Augmentation uses a network that generates new data by merging two or more samples from the same class~\cite{lemley2017smart}. Tran et al. generate augmented data, using a Bayesian approach, based on the distribution learned from the training set~\cite{tran2017bayesian}. DeVries and Taylor used simple transformations like noise, interpolations and extrapolations in the learned feature space to augment data~\cite{devries2017dataset}. Ratner et al., used generative adversarial networks to generate sequences of data augmentation operations~\cite{ratner2017learning}. More recently, several papers used the AutoAugment~\cite{cubuk2018autoaugment} search space with improved the optimization algorithms to find AutoAugment policies more efficiently~\cite{ho2019population, lim2019fast}.  

While all of the above approaches have worked on classification problems, we take an automated approach to finding optimal data augmentation policies for object detection. Unlike classification, labeled data for object detection is more scarce because it is more costly to annotate detection data. Compared to image classification, developing a data augmentation strategy for object detection is harder because there are more ways and complexities introduced by distorting the image, bounding box locations, and the sizes of the objects in detection datasets. Our goal is to use the validation set accuracy to help search for novel detection augmentation procedures using custom operations that generalize across datasets, dataset sizes, backbone architectures and detection algorithms.
\section{Methods}

 We treat data augmentation search as a discrete optimization problem and optimize for generalization performance.
 This work expands on previous work \cite{cubuk2018autoaugment} to focus on augmentation policies for object detection. Object detection introduces an additional complication of maintaining consistency between a bounding box location and a distorted image.
 Bounding box annotations open up the possibility of introducing augmentation operations that uniquely act upon the contents within each bounding box. Additionally, we explored how to change the bounding box locations when geometric transformations are applied to the image.

\begin{figure*}[t]
\centering
\includegraphics[width=0.85\linewidth]{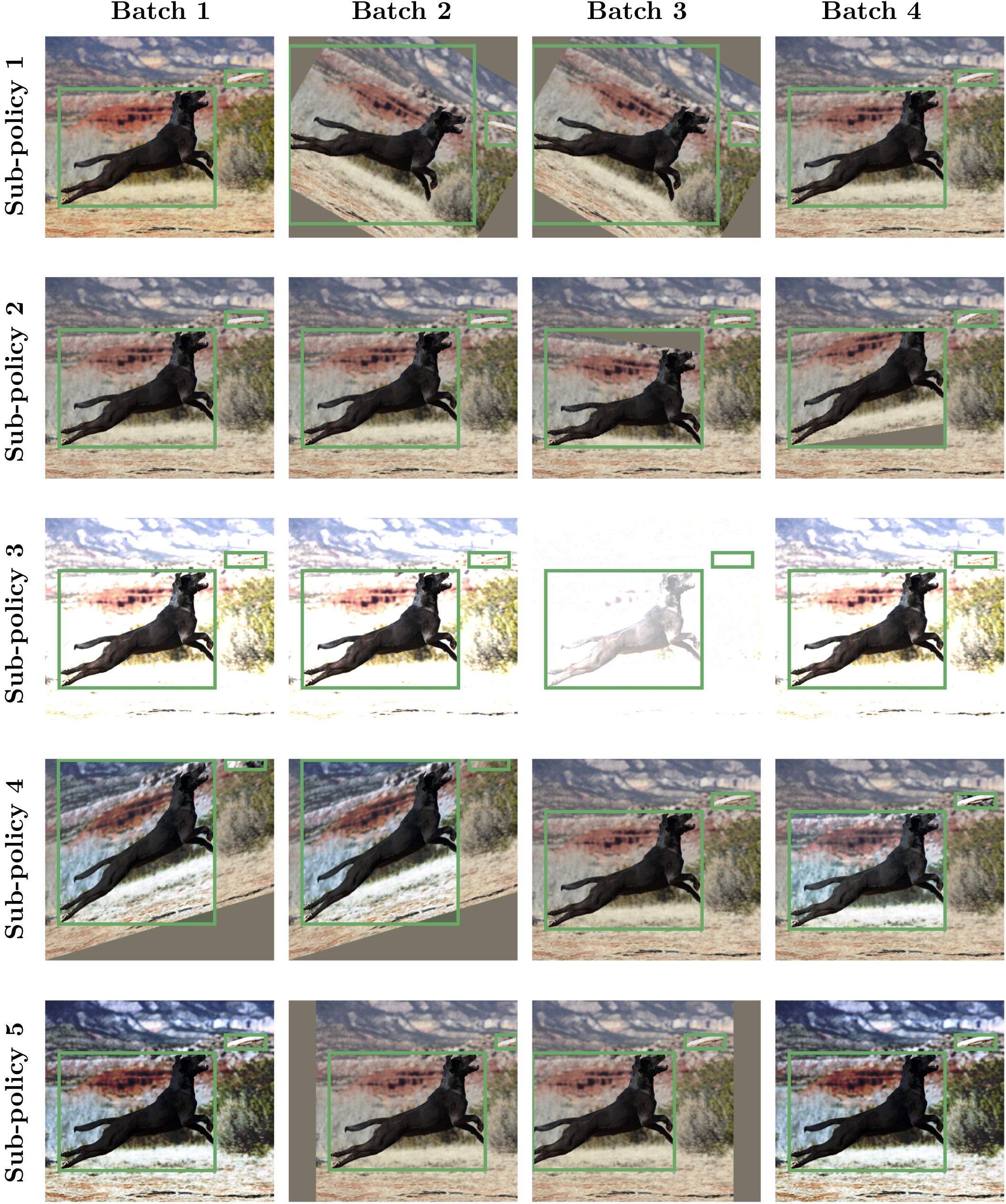}
\vspace{-0.2cm}
\small{
\begin{flushleft}
\hspace*{4.4cm} Sub-policy 1. \cmd{(Color, 0.2, 8), (Rotate, 0.8, 10)}\newline
\hspace*{4.4cm} Sub-policy 2. \cmd{(BBox\_Only\_ShearY, 0.8, 5)} \newline
\hspace*{4.4cm} Sub-policy 3. \cmd{(SolarizeAdd, 0.6, 8), (Brightness, 0.8, 10)} \newline
\hspace*{4.4cm} Sub-policy 4. \cmd{(ShearY, 0.6, 10), (BBox\_Only\_Equalize,0.6, 8)} \newline
\hspace*{4.4cm} Sub-policy 5. \cmd{(Equalize, 0.6, 10), (TranslateX, 0.2, 2)} \newline
\end{flushleft}
}
\vspace{-0.6cm}
\caption{\textbf{Examples of learned augmentation sub-policies.} 5 examples of learned sub-policies applied to one example image. Each column corresponds to a different random sample of the corresponding sub-policy. Each step of an augmentation sub-policy consists of a triplet corresponding to the operation, the probability of application and a magnitude measure. The bounding box is adjusted to maintain consistency with the applied augmentation. Note the probability and magnitude are discretized values (see text for details). }
\label{fig:example_strategy}
\end{figure*}

We define an augmentation policy as a unordered set of $K$ sub-policies. During training one of the $K$ sub-policies will be selected at random and then applied to the current image. Each sub-policy has $N$ image transformations which are applied sequentially. We turn this problem of searching for a learned augmentation policy into a discrete optimization problem by creating a search space ~\cite{cubuk2018autoaugment}.
The search space consists $K=5$ sub-policies with each sub-policy consisting of $N=2$ operations applied in sequence to a single image. Additionally, each operation is also associated with two hyperparameters specifying the probability of applying the operation, and the magnitude of the operation. Figure \ref{fig:example_strategy} (bottom text) demonstrates 5 of the learned sub-policies. 
The probability parameter introduces a notion of stochasticity into the augmentation policy whereby the selected augmentation operation will be applied to the image with the specified probability.

In several preliminary experiments, we identified 22 operations for the search space that appear beneficial for object detection. These operations were implemented in TensorFlow \cite{tensorflow-osdi2016}. We briefly summarize these operations, but reserve the details for the Appendix: 
\begin{itemize}
    \item \textbf{Color operations}. Distort color channels, without impacting the locations of the bounding boxes (e.g., \cmd{Equalize}, \cmd{Contrast}, \cmd{Brightness}). \footnote{The color transformations largely derive from transformation in the Python Image Library (PIL). \url{https://pillow.readthedocs.io/en/5.1.x/}} 
    
    \item \textbf{Geometric operations}. Geometrically distort the image, which correspondingly alters the location and size of the bounding box annotations (e.g., \cmd{Rotate}, \cmd{ShearX}, \cmd{TranslationY}, etc.).
    
    \item \textbf{Bounding box operations}. Only distort the pixel content contained within the bounding box annotations (e.g., \cmd{BBox\_Only\_Equalize}, \cmd{BBox\_Only\_Rotate}, \cmd{BBox\_Only\_FlipLR}).  
\end{itemize}

Note that for any operations that effected the geometry of an image, we likewise modified the bounding box size and location to maintain consistency.

We associate with each operation a custom range of parameter values and map this range on to a standardized range from 0 to 10.
We discretize the range of magnitude into $L$ uniformly-spaced values so that these parameters are amenable to discrete optimization. Similarly, we discretize the probability of applying an operation into $M$ uniformly-spaced values. 
In preliminary experiments we found that setting $L=6$ and $M=6$ provide a good balance between computational tractability and learning performance with an RL algorithm.
Thus, finding a good sub-policy becomes a search in a discrete space containing a cardinality of $(22LM)^2$.
In particular, to search over 5 sub-policies, the search space contains roughly $(22 \times 6 \times 6)^{2\times5} \approx 9.6\times 10^{28}$ possibilities and requires an efficient search technique to navigate this space.

Many methods exist for addressing the discrete optimization problem including reinforcement learning \cite{zoph2016neural}, evolutionary methods \cite{real2018regularized} and sequential model-based optimization \cite{liu2017progressive}. In this work, we choose to build on previous work by structuring the discrete optimization problem as the output space of an RNN and employ reinforcement learning to update the weights of the model \cite{zoph2016neural}. The training setup for the RNN is similar to~\cite{zoph2016neural,zoph2017learning, cubuk2017intriguing, cubuk2018autoaugment}. We employ the proximal policy optimization (PPO) \cite{schulman2017proximal} for the search algorithm. The RNN is unrolled 30 steps to predict a single augmentation policy. The number of unrolled steps, 30,  corresponds to the number of discrete predictions that must be made in order to enumerate 5 sub-policies. Each sub-policy consists of 2 operations and each operation consists of 3 predictions corresponding to the selected image transformation, probability of application and magnitude of the transformation.

In order to train each child model, we selected 5K images from the COCO training set as we found that searching directly on the full COCO dataset to be prohibitively expensive. We found that policies identified with this subset of data generalize to the full dataset while providing significant computational savings. Briefly, we trained each child model\footnote{We employed a base learning rate of 0.08 over 150 epochs; image size was $640 \times 640$; $\alpha=0.25$ and $\gamma=1.5$ for the focal loss parameters; weight decay of $1e-4$; batch size was 64} from scratch on the 5K COCO images with the ResNet-50 backbone \cite{he2016deep} and RetinaNet detector \cite{lin2017focal} using a cosine learning rate decay \cite{cosine}.
The reward signal for the controller is the mAP on a custom held-out validation set of 7392 images created from a subset of the COCO training set.

The RNN controller is trained over 20K augmentation policies. The search employed 400 TPU's~\cite{jouppi2017datacenter} over 48 hours with identical hyper-parameters for the controller as \cite{zoph2017learning}. The search can be sped up using the recently developed, more efficient search methods based on population based training~\cite{ho2019population} or density matching~\cite{lim2019fast}. The learned policy can be seen in Table~\ref{tab:policy} in the Appendix.

\section{Results}
\label{sec:results}

We applied our automated augmentation method on the COCO dataset with a ResNet-50~\cite{he2016deep} backbone with RetinaNet~\cite{lin2017focal} in order to find good augmentation policies to generalize to other detection datasets. We  use the top policy found on COCO and apply it to different datasets, dataset sizes and architecture configurations to examine generalizability and how the policy fares in a limited data regime. 

\subsection{Learning a data augmentation policy}

Searching for the learned augmentation strategy on 5K COCO training images resulted in the final augmentation policy that will be used in all of our results. Upon inspection, the most commonly used operation in good policies is \cmd{Rotate}, which rotates the whole image and the bounding boxes. The bounding boxes end up larger after the rotation, to include all of the rotated object. Despite this effect of the \cmd{Rotate} operation, it seems to be very beneficial: it is the most frequently used operation in good policies. Two other operations that are commonly used are \cmd{Equalize} and \cmd{BBox\_Only\_TranslateY}. \cmd{Equalize} flattens the histogram of the pixel values, and does not modify the location or size of each bounding box. \cmd{BBox\_Only\_TranslateY} translates only the objects in bounding boxes vertically, up or down with equal probability.

\subsection{Learned augmentation policy systematically improves object detection}
We assess the quality of the top augmentation policy on the competitive COCO dataset \cite{lin2014microsoft} on different backbone architectures and detection algorithms. We start with the competitive RetinaNet object detector
\footnote{{\texttt{\href{}{https://github.com/tensorflow/tpu}}}}
employing the same training protocol as \cite{ghiasi2018dropblock}. Briefly, we train from scratch with a global batch size of 64, images were resized to $640 \times 640$, learning rate of 0.08, weight decay of $1e-4$, $\alpha=0.25$ and $\gamma=1.5$ for the focal loss parameters, trained for 150 epochs, used stepwise decay where the learning rate was reduced by a factor of 10 at epochs 120 and 140.
All models were trained on TPUs ~\cite{jouppi2017datacenter}.

The baseline RetinaNet architecture used in this and subsequent sections employs standard data augmentation techniques largely tailored to image classification training \cite{lin2017focal}. This consists of doing horizontal flipping with 50\% probability and multi-scale jittering where images are randomly resized between 512 and 786 during training and then cropped to 640x640. 

Our results using our augmentation policy on the above procedures are shown in Tables \ref{tab:across_models} and \ref{tab:search_space}. In Table \ref{tab:across_models} the learned augmentation policy achieves systematic gains across a several backbone architectures with improvements ranging from +1.6 mAP to +2.3 mAP.
In comparison, a previous state-of-the-art regularization technique applied to ResNet-50 \cite{ghiasi2018dropblock} achieves a gain of +1.7\% mAP (Table \ref{tab:search_space}).

To better understand where the gains come from, we break the data augmentation strategies applied to ResNet-50 into three parts: color operations, geometric operations, and bbox-only-operations (Table \ref{tab:search_space}). Employing color operations only boosts performance by +0.8 mAP.
Combining the search with geometric operations increases the boost in performance by +1.9 mAP.
Finally, adding bounding box-specific operations
yields the best results when used in conjunction with the previous operations and provides +2.3\% mAP improvement over the baseline. Note that the policy found was only searched using 5K COCO training examples and still generalizes extremely well when trained on the full COCO dataset.

\begin{table}[t]
\centering
\begin{tabular}{lrrr}
  \hline
  Backbone & Baseline & Our result & Difference  \\
  \hline
  ResNet-50 &  36.7 & 39.0 & +2.3 \\
  ResNet-101 &  38.8 & 40.4 & +1.6 \\
  ResNet-200 &  39.9 & 42.1 & +2.2 \\
  \hline
\end{tabular}
\caption{\textbf{Improvements with learned augmentation policy across different ResNet backbones.} All results employ RetinaNet detector \cite{lin2017focal} on the COCO dataset \cite{lin2014microsoft}.}
\label{tab:across_models}  
\end{table}

\begin{table}[t]
\centering
\begin{tabular}{l|c}
  \hline
  Method & mAP  \\
  \hline
  baseline & 36.7 \\
  \hline
  baseline + DropBlock~\cite{ghiasi2018dropblock} & 38.4 \\
  \hline 
  Augmentation policy with color operations &  37.5\\
  + geometric operations & 38.6 \\
  + bbox-only operations & \textbf{39.0} \\
  \hline
\end{tabular}
\caption{\textbf{Improvements in object detection with learned augmentation policy.} All results employ RetinaNet detector with ResNet-50 backbone \cite{lin2017focal} on COCO dataset \cite{lin2014microsoft}. DropBlock shows gain in performance employing a state-of-the-art regularization method \cite{ghiasi2018dropblock}.}
\label{tab:search_space}  
\end{table}

\subsection{Exploiting learned augmentation policies achieves state-of-the-art object detection}
A good data augmentation policy is one that can transfer between models, between datasets and work well for models trained on different image sizes. Here we experiment with the learned augmentation policy
on a different backbone architecture and detection model.
To test how the learned policy transfers to a state-of-the-art detection model, we replace the ResNet-50 backbone with the AmoebaNet-D architecture~\cite{real2018regularized}. The detection algorithm was changed from RetinaNet~\cite{lin2017focal} to NAS-FPN~\cite{ghiasi2019NASFpn}. Additionally, we use ImageNet pre-training for the AmoebaNet-D backbone as we found we are not able to achieve competitive results when training from scratch. The model was trained for 150 epochs using a cosine learning rate decay with a learning rate of 0.08. The rest of the setup was identical to the ResNet-50 backbone model except the image size was increased from $640 \times 640$ to $1280 \times 1280$. 

Table~\ref{tab:large_results} indicates that the learned augmentation policy improves
+1.5\% mAP on top of a competitive, detection architecture and setup. These experiments additionally show that the augmentation policy transfers well across a different backbone architecture, detection algorithm, image sizes (i.e. $640 \rightarrow 1280$ pixels), and training procedure (training from scratch $\rightarrow$ using ImageNet pre-training) .
We can extend these results even further by increasing the image resolution from $1280$ to $1536$ pixels and likewise increasing the number of detection anchors\footnote{Specifically, we increase the number of anchors from $3\times 3$ to $9 \times 9$ by changing the aspect ratios from \{1/2, 1, 2\} to \{1/5, 1/4, 1/3, 1/2, 1, 2, 3, 4, 5\}. When making this change we increased the strictness in the IoU thresholding from 0.5/0.5 to 0.6/0.5 due to the increased number of anchors following \cite{yang2018metaanchor}. The anchor scale was also increased from 4 to 5 to compensate for the larger image size.} following \cite{yang2018metaanchor}. Since this model is significantly larger than the previous models, we increase the number of sub-policies in the learned policy by combining the top 4 policies from the search, which leads to a 20 sub-policy learned augmentation.

\begin{table*}[t]
\centering
\small
\begin{tabular}{r|c|c|c|ccc}
  Architecture  & Change & \# Scales & mAP & mAP$_{\texttt S}$ & mAP$_{\texttt M}$ & mAP$_{\texttt L}$ \\
  \toprule
  MegDet \cite{Peng_2018_CVPR} & & multiple & 50.5 & - & - & - \\
  \hline
  \multirow{3}{*}{AmoebaNet + NAS-FPN}  & baseline \cite{ghiasi2019NASFpn} & 1 & 47.0 & 30.6 & 50.9 & 61.3 \\
   & + learned augmentation & 1 & 48.6 & 32.0 & 53.4 & 62.7 \\
   & + $\uparrow$ anchors, $\uparrow$ image size & 1 & \textbf{50.7} & \textbf{34.2} & \textbf{55.5} & \textbf{64.5} \\[0.1cm]
\end{tabular}
\caption{\textbf{Exceeding state-of-the-art detection with learned augmentation policy.} Reporting mAP for COCO validation set. Previous state-of-the-art results for COCO detection evaluated a single image at multiple spatial scales to perform detection at test time~\cite{Peng_2018_CVPR}. Our current results only require a single inference computation at single spatial scale.
Backbone model is AmoebaNet-D \cite{real2018regularized} and the NAS-FPN detection system \cite{ghiasi2019NASFpn}. For the \textbf{50.7} result, in addition to using the learned data augmentation policy, we increase the image size from 1280 to 1536 and the number of detection anchors from 3x3 to 9x9.}
\label{tab:large_results}  
\end{table*}

This result of these simple modifications is the first single-stage detection system to achieve state-of-the-art, single-model results of 50.7 mAP on COCO. We note that this result only requires a single pass of the image, where as the previous results required multiple evaluations of the same image at different spatial scales at test time \cite{Peng_2018_CVPR}. Additionally, these results were arrived at by increasing the image resolution and increasing the number of anchors - both simple and well known techniques for improving object detection performance \cite{yang2018metaanchor, huang2017speed}. In contrast, previous state-of-the-art results relied on roughly multiple, custom modifications of the model architecture and regularization methods in order to achieve these results \cite{Peng_2018_CVPR}. Our method largely relies on a more modern network architecture paired with a learned data augmentation policy.

\subsection{Learned augmentation policies transfer to other detection datasets.}

To evaluate the transferability of the learned policies to an entirely different dataset and another different detection algorithm, we train a Faster R-CNN~\cite{ren2015faster} model with a ResNet-101 backbone on PASCAL VOC dataset~\cite{everingham2010pascal}. We combine the training sets of PASCAL VOC 2007 and PASCAL VOC 2012, and test our model on the PASCAL VOC 2007 test set (4952 images). Our evaluation metric is the mean average precision at an IoU threshold of 0.5 (mAP50). For the baseline model, we use the Tensorflow Object Detection API~\cite{huang2017speed} with the default hyperparameters: 9 GPU workers are utilized for asynchronous training where each worker processes a batch size of 1. Initial learning rate is set to be $3\times 10^{-4}$, which is decayed by 0.1 after 500K steps. Training is started from a COCO detection model checkpoint. When training with our data augmentation policy, we do not change any of the training details, and just add our policy found on COCO to the pre-processing. This leads to a 2.7\% improvement on mAP50 (Table~\ref{tab:pascal_results}). 

\begingroup
\setlength{\tabcolsep}{2pt}
\begin{table*}[t]
\centering
\small
\begin{tabular}{l|llllllllllllllllllll|c}

&plane&bike&bird&boat&bottle&bus&car&cat&chair&cow&table&dog&horse&mbike&person&plant&sheep&sofa&train&tv&\textbf{mean}\\
\hline
baseline&86.6&82.2&75.9&63.4&62.3&84.7&86.8&92.0&55.5&83.3&63.1&89.2&89.4&85.0&85.6&50.7&76.2&73.0&86.6&76.3&76.0\\
\hline
ours&88.0&83.3&78.0&65.9&63.5&85.5&87.4&93.1&58.5&83.9&65.2&90.1&90.2&85.9&86.6&55.2&78.6&76.6&88.6&80.3&78.7\\
\hline
\end{tabular}
\caption{\textbf{Learned augmentation policy transfer to other object detection tasks.} Mean average precision (\%) at IoU threshold 0.5 on a Faster R-CNN detector \cite{ren2015faster} with a ResNet-101 backbone trained and evaluated on PASCAL VOC 2007 \cite{everingham2010pascal}. Note that the augmentation policy was learned from the policy search on the COCO dataset.}
\label{tab:pascal_results}  
\end{table*}
\endgroup

\subsection{Learned augmentation policies mimic the performance of larger annotated datasets}

In this section we conducted experiments to determine how the learned augmentation policy will perform if there is more or less training data. To conduct these experiments we took subsets of the COCO dataset to make datasets with the following number of images: 5000, 9000, 14000, 23000 (see Table~\ref{tab:reduced_coco}). All models trained in this experiment are using a ResNet-50 backbone with RetinaNet and are trained for 150 epochs without using ImageNet pretraining.

\begin{table*}[h!]
\centering
\small
\begin{tabular}{c|rrrr|rrrr}
  \hline
  training & \multicolumn{4}{c|}{Baseline} & \multicolumn{4}{c}{Our results}  \\
   set size & mAP$_{\texttt S}$ &  mAP$_{\texttt M}$ &  mAP$_{\texttt L}$ & mAP & mAP$_{\texttt S}$ &  mAP$_{\texttt M}$ &  mAP$_{\texttt L}$ & mAP \\
  \hline
  5000  &1.9 &7.1& 9.7& 6.5 &3.2 &9.8&12.7& 8.7 \\
  9000 &4.3&12.3& 17.6& 11.8  &7.1 & 16.8&22.3& 15.1 \\
  14000 &6.8&17.5& 23.9& 16.4 &9.5 & 22.1&29.8& 19.9 \\
  23000 &10.0&24.3& 33.3& 22.6 & 11.9&27.8&36.8& 25.3  \\
  \hline
\end{tabular}
\caption{\textbf{Learned augmentation policy is especially beneficial for small datasets and small objects.} Mean average precision (mAP) for RetinaNet model trained on COCO with varying subsets of the original training set. mAP$_{\texttt S}$, mAP$_{\texttt M}$ and mAP$_{\texttt L}$ denote the mean average precision for small, medium and large examples. Note the complete COCO training set consists of 118K examples.The same policy found on the 5000 COCO images was used in all of the experiments. The models in the first row were trained on the same 5000 images that the policies were searched on. }
\label{tab:reduced_coco}  
\end{table*}

\begin{figure}[h!]
\centering
\includegraphics[width=1.0\linewidth]{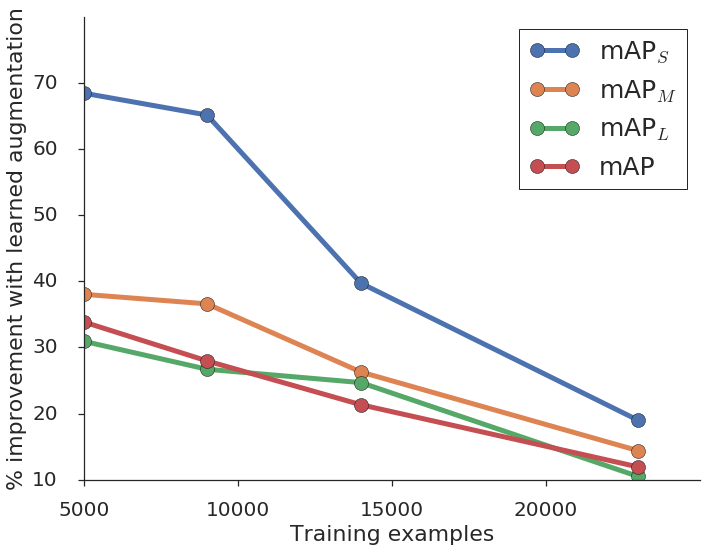}
\caption{Percentage improvement in mAP for objects of different sizes due to the learned augmentation policy.}
\label{fig:percentage_improvement}
\end{figure}

As we expected, the improvements due to the learned augmentation policy is larger when the model is trained on smaller datasets, which can be seen in Fig.~\ref{fig:percentage_improvement} and in Table~\ref{tab:reduced_coco}. We show that for models trained on 5,000 training samples, the learned augmentation policy can improve mAP by more than 70\% relative to the baseline. As the training set size is increased, the effect of the learned augmentation policy is decreased, although the improvements are still significant. It is interesting to note that models trained with learned augmentation policy seem to do especially well on detecting smaller objects, especially when fewer images are present in the training dataset. For example, for small objects, applying the learned augmentation policy seems to be better than increasing the dataset size by 50\%, as seen in Table.~\ref{tab:reduced_coco}. For small objects, training with the learned augmentation policy with 9000 examples results in better performance than the baseline when using 15000 images. In this scenario using our augmentation policy is almost as effective as doubling your dataset size.

\begin{figure}[h!]
\centering
\includegraphics[width=1.0\linewidth]{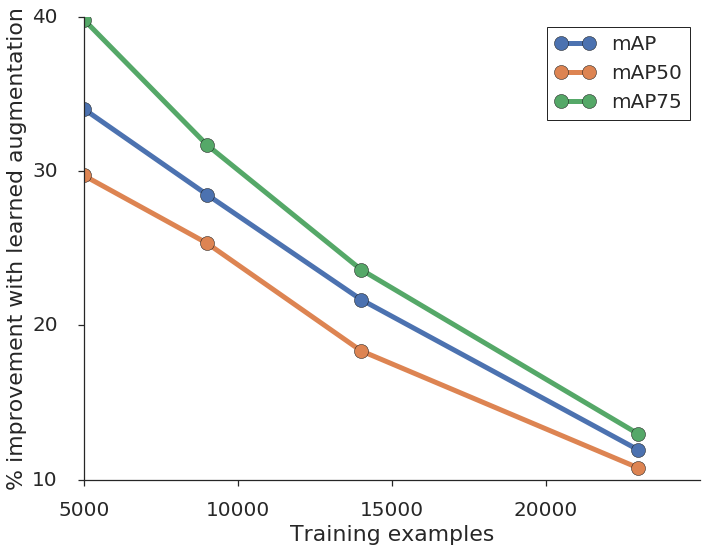}
\caption{Percentage improvement due to the learned augmentation policy on mAP, AP50, and AP75, relative to models trained with baseline augmentation.  
}
\label{fig:AP75}
\end{figure}

Another interesting behavior of models trained with the learned augmentation policy is that they do relatively better on the harder task of AP75 (average precision IoU=0.75). In Fig.~\ref{fig:AP75}, we plot the percentage improvement in mAP, AP50, and AP75 for models trained with the learned augmentation policy (relative to baseline augmentation). The relative improvement of AP75 is larger than that of AP50 for all training set sizes. The learned data augmentation is particularly beneficial at AP75 indicating that the augmentation policy helps with more precisely aligning the bounding box prediction. This suggests that the augmentation policy particularly helps with learned fine spatial details in bounding box position -- which is consistent with the gains observed with small objects.

\subsection{Learned data augmentation improves model regularization}
\begin{figure}[h!]
\centering
\includegraphics[width=1.0\linewidth]{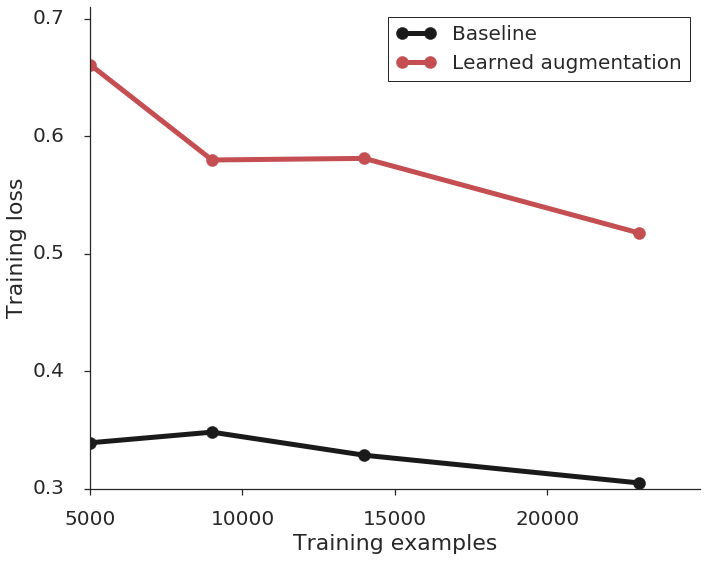}
\caption{Training loss vs. number of training examples for baseline model (black) and with the learned augmentation policy (red).  
}
\label{fig:training_loss}
\end{figure}

\begin{figure}[h!]
\centering
\includegraphics[width=1.0\linewidth]{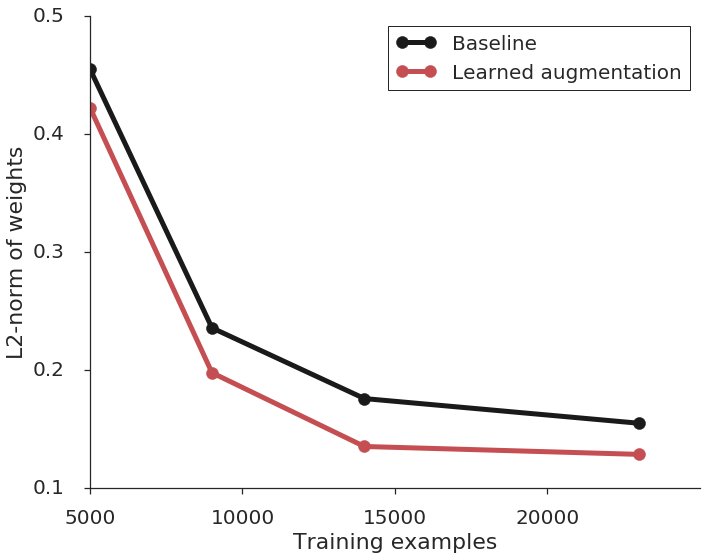}
\caption{L$_2$ norm of the weights of the baseline (black) and our (red) models at the end of training. Note that the L$_2$ norm of the weights decrease with increasing training set size. The learned augmentation policy further decreases the norm of the weights.  
}
\label{fig:l2_norm}
\end{figure}

In this section, we study the regularization effect of the learned data augmentation. We first notice that the final training loss of a detection models is lower when trained on a larger training set (see black curve in Fig.~\ref{fig:training_loss}). When we apply the learned data augmentation, the training loss is increased significantly for all dataset sizes (red curve). The  regularization effect can also be seen by looking at the L$_2$ norm of the weights of the trained models. The L$_2$ norm of the weights is smaller for models trained on larger datasets, and models trained with the learned augmentation policy have a smaller L$_2$ norm than models trained with baseline augmentation (see Fig.~\ref{fig:l2_norm}). 

\section{Discussion}
In this work, we investigate the application of a learned data augmentation policy on object detection performance. We find that a learned data augmentation policy is effective across all data sizes considered, with a larger improvement when the training set is small. We also observe that the improvement due to a learned data augmentation policy is larger on harder tasks of detecting smaller objects and detecting with more precision. 

We also find that other successful regularization techniques are not beneficial when applied in tandem with a learned data augmentation policy. We carried out several experiments with Input Mixup~\cite{zhang2017mixup}, Manifold Mixup~\cite{verma2018manifold} and Dropblock~\cite{ghiasi2018dropblock}. For all methods we found that they either did not help nor hurt model performance. This is an interesting result as the proposed method independently outperforms these regularization methods, yet apparently these regularization methods are not needed when applying a learned data augmentation policy.

Future work will include the application of this method to other perceptual domains. For example, a natural extension of a learned augmentation policy would be to semantic~\cite{long2015fully} and instance segmentation~\cite{pinheiro2016learning,dai2016instance}. Likewise, point cloud featurizations ~\cite{qi2017pointnet,qi2017pointnet++} are another domain that has a rich set of possibilities for geometric data augmentation operations, and can benefit from an approach similar to the one taken here. Human annotations required for acquiring training set examples for such tasks are costly. Based on our findings, learned augmentation policies are transferable and are more effective for models trained on limited training data. Thus, investing in libraries for learning data augmentation policies may be an efficient alternative to acquiring additional human annotated data.  

\section*{Acknowledgments}
We thank Ruoming Pang and the rest of the Brain team for their help.



\begin{thebibliography}{10}\itemsep=-1pt

\bibitem{tensorflow-osdi2016}
M.~Abadi, P.~Barham, J.~Chen, Z.~Chen, A.~Davis, J.~Dean, M.~Devin,
  S.~Ghemawat, G.~Irving, M.~Isard, M.~Kudlur, J.~Levenberg, R.~Monga,
  S.~Moore, D.~G. Murray, B.~Steiner, P.~Tucker, V.~Vasudevan, P.~Warden,
  M.~Wicke, Y.~Yu, and X.~Zheng.
\newblock Tensorflow: A system for large-scale machine learning.
\newblock In {\em Proceedings of the 12th USENIX Conference on Operating
  Systems Design and Implementation}, OSDI'16, pages 265--283, Berkeley, CA,
  USA, 2016. USENIX Association.

\bibitem{antoniou2017data}
A.~Antoniou, A.~Storkey, and H.~Edwards.
\newblock Data augmentation generative adversarial networks.
\newblock {\em arXiv preprint arXiv:1711.04340}, 2017.

\bibitem{baird1992document}
H.~S. Baird.
\newblock Document image defect models.
\newblock In {\em Structured Document Image Analysis}, pages 546--556.
  Springer, 1992.

\bibitem{ciregan2012multi}
D.~Ciregan, U.~Meier, and J.~Schmidhuber.
\newblock Multi-column deep neural networks for image classification.
\newblock In {\em Proceedings of IEEE Conference on Computer Vision and Pattern
  Recognition}, pages 3642--3649. IEEE, 2012.

\bibitem{cubuk2018autoaugment}
E.~D. Cubuk, B.~Zoph, D.~Mane, V.~Vasudevan, and Q.~V. Le.
\newblock Autoaugment: Learning augmentation policies from data.
\newblock {\em arXiv preprint arXiv:1805.09501}, 2018.

\bibitem{cubuk2017intriguing}
E.~D. Cubuk, B.~Zoph, S.~S. Schoenholz, and Q.~V. Le.
\newblock Intriguing properties of adversarial examples.
\newblock {\em arXiv preprint arXiv:1711.02846}, 2017.

\bibitem{dai2016instance}
J.~Dai, K.~He, and J.~Sun.
\newblock Instance-aware semantic segmentation via multi-task network cascades.
\newblock In {\em Proceedings of the IEEE Conference on Computer Vision and
  Pattern Recognition}, pages 3150--3158, 2016.

\bibitem{devries2017dataset}
T.~DeVries and G.~W. Taylor.
\newblock Dataset augmentation in feature space.
\newblock {\em arXiv preprint arXiv:1702.05538}, 2017.

\bibitem{cutout2017}
T.~DeVries and G.~W. Taylor.
\newblock Improved regularization of convolutional neural networks with cutout.
\newblock {\em arXiv preprint arXiv:1708.04552}, 2017.

\bibitem{dwibedi2017cut}
D.~Dwibedi, I.~Misra, and M.~Hebert.
\newblock Cut, paste and learn: Surprisingly easy synthesis for instance
  detection.
\newblock In {\em Proceedings of the IEEE International Conference on Computer
  Vision}, pages 1301--1310, 2017.

\bibitem{everingham2010pascal}
M.~Everingham, L.~Van~Gool, C.~K. Williams, J.~Winn, and A.~Zisserman.
\newblock The pascal visual object classes (voc) challenge.
\newblock {\em International journal of computer vision}, 88(2):303--338, 2010.

\bibitem{ford2019adversarial}
N.~Ford, J.~Gilmer, N.~Carlini, and D.~Cubuk.
\newblock Adversarial examples are a natural consequence of test error in
  noise.
\newblock {\em arXiv preprint arXiv:1901.10513}, 2019.

\bibitem{ghiasi2018dropblock}
G.~Ghiasi, T.-Y. Lin, and Q.~V. Le.
\newblock {DropBlock}: A regularization method for convolutional networks.
\newblock In {\em Advances in Neural Information Processing Systems}, pages
  10750--10760, 2018.

\bibitem{ghiasi2019NASFpn}
G.~Ghiasi, T.-Y. Lin, R.~Pang, and Q.~V. Le.
\newblock {NAS-FPN}: Learning scalable feature pyramid architecture for object
  detection.
\newblock In {\em The IEEE Conference on Computer Vision and Pattern
  Recognition (CVPR)}, June 2019.

\bibitem{girshick2018detectron}
R.~Girshick, I.~Radosavovic, G.~Gkioxari, P.~Doll{\'a}r, and K.~He.
\newblock Detectron, 2018.

\bibitem{he2016deep}
K.~He, X.~Zhang, S.~Ren, and J.~Sun.
\newblock Deep residual learning for image recognition.
\newblock In {\em Proceedings of the IEEE Conference on Computer Vision and
  Pattern Recognition (CVPR)}, pages 770--778, 2016.

\bibitem{ho2019population}
D.~Ho, E.~Liang, I.~Stoica, P.~Abbeel, and X.~Chen.
\newblock Population based augmentation: Efficient learning of augmentation
  policy schedules.
\newblock {\em arXiv preprint arXiv:1905.05393}, 2019.

\bibitem{hu2017squeeze}
J.~Hu, L.~Shen, and G.~Sun.
\newblock Squeeze-and-excitation networks.
\newblock {\em arXiv preprint arXiv:1709.01507}, 2017.

\bibitem{huang2017speed}
J.~Huang, V.~Rathod, C.~Sun, M.~Zhu, A.~Korattikara, A.~Fathi, I.~Fischer,
  Z.~Wojna, Y.~Song, S.~Guadarrama, et~al.
\newblock Speed/accuracy trade-offs for modern convolutional object detectors.
\newblock In {\em Proceedings of the IEEE conference on computer vision and
  pattern recognition}, pages 7310--7311, 2017.

\bibitem{jouppi2017datacenter}
N.~P. Jouppi, C.~Young, N.~Patil, D.~Patterson, G.~Agrawal, R.~Bajwa, S.~Bates,
  S.~Bhatia, N.~Boden, A.~Borchers, et~al.
\newblock In-datacenter performance analysis of a tensor processing unit.
\newblock In {\em 2017 ACM/IEEE 44th Annual International Symposium on Computer
  Architecture (ISCA)}, pages 1--12. IEEE, 2017.

\bibitem{krizhevsky2012imagenet}
A.~Krizhevsky, I.~Sutskever, and G.~E. Hinton.
\newblock Imagenet classification with deep convolutional neural networks.
\newblock In {\em Advances in Neural Information Processing Systems}, 2012.

\bibitem{lemley2017smart}
J.~Lemley, S.~Bazrafkan, and P.~Corcoran.
\newblock Smart augmentation learning an optimal data augmentation strategy.
\newblock {\em IEEE Access}, 5:5858--5869, 2017.

\bibitem{lim2019fast}
S.~Lim, I.~Kim, T.~Kim, C.~Kim, and S.~Kim.
\newblock Fast autoaugment.
\newblock {\em arXiv preprint arXiv:1905.00397}, 2019.

\bibitem{lin2017focal}
T.-Y. Lin, P.~Goyal, R.~Girshick, K.~He, and P.~Doll{\'a}r.
\newblock Focal loss for dense object detection.
\newblock In {\em Proceedings of the IEEE international conference on computer
  vision}, pages 2980--2988, 2017.

\bibitem{lin2014microsoft}
T.-Y. Lin, M.~Maire, S.~Belongie, J.~Hays, P.~Perona, D.~Ramanan,
  P.~Doll{\'a}r, and C.~L. Zitnick.
\newblock Microsoft coco: Common objects in context.
\newblock In {\em European conference on computer vision}, pages 740--755.
  Springer, 2014.

\bibitem{liu2017progressive}
C.~Liu, B.~Zoph, J.~Shlens, W.~Hua, L.-J. Li, L.~Fei-Fei, A.~Yuille, J.~Huang,
  and K.~Murphy.
\newblock Progressive neural architecture search.
\newblock {\em arXiv preprint arXiv:1712.00559}, 2017.

\bibitem{liu2016ssd}
W.~Liu, D.~Anguelov, D.~Erhan, C.~Szegedy, S.~Reed, C.-Y. Fu, and A.~C. Berg.
\newblock Ssd: Single shot multibox detector.
\newblock In {\em European conference on computer vision}, pages 21--37.
  Springer, 2016.

\bibitem{long2015fully}
J.~Long, E.~Shelhamer, and T.~Darrell.
\newblock Fully convolutional networks for semantic segmentation.
\newblock In {\em Proceedings of the IEEE conference on computer vision and
  pattern recognition}, pages 3431--3440, 2015.

\bibitem{lopes2019improving}
R.~G. Lopes, D.~Yin, B.~Poole, J.~Gilmer, and E.~D. Cubuk.
\newblock Improving robustness without sacrificing accuracy with patch gaussian
  augmentation.
\newblock {\em arXiv preprint arXiv:1906.02611}, 2019.

\bibitem{cosine}
I.~Loshchilov and F.~Hutter.
\newblock {SGDR}: Stochastic gradient descent with warm restarts.
\newblock {\em arXiv preprint arXiv:1608.03983}, 2016.

\bibitem{mun2017generative}
S.~Mun, S.~Park, D.~K. Han, and H.~Ko.
\newblock Generative adversarial network based acoustic scene training set
  augmentation and selection using svm hyper-plane.
\newblock In {\em Detection and Classification of Acoustic Scenes and Events
  Workshop}, 2017.

\bibitem{Peng_2018_CVPR}
C.~Peng, T.~Xiao, Z.~Li, Y.~Jiang, X.~Zhang, K.~Jia, G.~Yu, and J.~Sun.
\newblock Megdet: A large mini-batch object detector.
\newblock In {\em The IEEE Conference on Computer Vision and Pattern
  Recognition (CVPR)}, June 2018.

\bibitem{perez2017effectiveness}
L.~Perez and J.~Wang.
\newblock The effectiveness of data augmentation in image classification using
  deep learning.
\newblock {\em arXiv preprint arXiv:1712.04621}, 2017.

\bibitem{pinheiro2016learning}
P.~O. Pinheiro, T.-Y. Lin, R.~Collobert, and P.~Doll{\'a}r.
\newblock Learning to refine object segments.
\newblock In {\em European Conference on Computer Vision}, pages 75--91.
  Springer, 2016.

\bibitem{qi2017pointnet}
C.~R. Qi, H.~Su, K.~Mo, and L.~J. Guibas.
\newblock Pointnet: Deep learning on point sets for 3d classification and
  segmentation.
\newblock In {\em Proceedings of the IEEE Conference on Computer Vision and
  Pattern Recognition}, pages 652--660, 2017.

\bibitem{qi2017pointnet++}
C.~R. Qi, L.~Yi, H.~Su, and L.~J. Guibas.
\newblock Pointnet++: Deep hierarchical feature learning on point sets in a
  metric space.
\newblock In {\em Advances in Neural Information Processing Systems}, pages
  5099--5108, 2017.

\bibitem{ratner2017learning}
A.~J. Ratner, H.~Ehrenberg, Z.~Hussain, J.~Dunnmon, and C.~R{\'e}.
\newblock Learning to compose domain-specific transformations for data
  augmentation.
\newblock In {\em Advances in Neural Information Processing Systems}, pages
  3239--3249, 2017.

\bibitem{real2018regularized}
E.~Real, A.~Aggarwal, Y.~Huang, and Q.~V. Le.
\newblock Regularized evolution for image classifier architecture search.
\newblock In {\em Thirty-Third AAAI Conference on Artificial Intelligence},
  2019.

\bibitem{ren2015faster}
S.~Ren, K.~He, R.~Girshick, and J.~Sun.
\newblock Faster r-cnn: Towards real-time object detection with region proposal
  networks.
\newblock In {\em Advances in neural information processing systems}, pages
  91--99, 2015.

\bibitem{sato2015apac}
I.~Sato, H.~Nishimura, and K.~Yokoi.
\newblock Apac: Augmented pattern classification with neural networks.
\newblock {\em arXiv preprint arXiv:1505.03229}, 2015.

\bibitem{schulman2017proximal}
J.~Schulman, F.~Wolski, P.~Dhariwal, A.~Radford, and O.~Klimov.
\newblock Proximal policy optimization algorithms.
\newblock {\em arXiv preprint arXiv:1707.06347}, 2017.

\bibitem{simard2003best}
P.~Y. Simard, D.~Steinkraus, J.~C. Platt, et~al.
\newblock Best practices for convolutional neural networks applied to visual
  document analysis.
\newblock In {\em Proceedings of International Conference on Document Analysis
  and Recognition}, 2003.

\bibitem{sixt2016rendergan}
L.~Sixt, B.~Wild, and T.~Landgraf.
\newblock Rendergan: Generating realistic labeled data.
\newblock {\em arXiv preprint arXiv:1611.01331}, 2016.

\bibitem{szegedy2015going}
C.~Szegedy, W.~Liu, Y.~Jia, P.~Sermanet, S.~Reed, D.~Anguelov, D.~Erhan,
  V.~Vanhoucke, A.~Rabinovich, et~al.
\newblock Going deeper with convolutions.
\newblock In {\em Proceedings of the IEEE Conference on Computer Vision and
  Pattern Recognition (CVPR)}, 2015.

\bibitem{tran2017bayesian}
T.~Tran, T.~Pham, G.~Carneiro, L.~Palmer, and I.~Reid.
\newblock A bayesian data augmentation approach for learning deep models.
\newblock In {\em Advances in Neural Information Processing Systems}, pages
  2794--2803, 2017.

\bibitem{verma2018manifold}
V.~Verma, A.~Lamb, C.~Beckham, A.~Courville, I.~Mitliagkis, and Y.~Bengio.
\newblock Manifold mixup: Encouraging meaningful on-manifold interpolation as a
  regularizer.
\newblock {\em arXiv preprint arXiv:1806.05236}, 2018.

\bibitem{wan2013regularization}
L.~Wan, M.~Zeiler, S.~Zhang, Y.~Le~Cun, and R.~Fergus.
\newblock Regularization of neural networks using dropconnect.
\newblock In {\em International Conference on Machine Learning}, pages
  1058--1066, 2013.

\bibitem{wang2017fast}
X.~Wang, A.~Shrivastava, and A.~Gupta.
\newblock A-fast-rcnn: Hard positive generation via adversary for object
  detection.
\newblock In {\em Proceedings of the IEEE Conference on Computer Vision and
  Pattern Recognition}, pages 2606--2615, 2017.

\bibitem{yang2018metaanchor}
T.~Yang, X.~Zhang, Z.~Li, W.~Zhang, and J.~Sun.
\newblock Metaanchor: Learning to detect objects with customized anchors.
\newblock In {\em Advances in Neural Information Processing Systems}, pages
  318--328, 2018.

\bibitem{yin2019afourier}
D.~Yin, R.~G. Lopes, J.~Shlens, E.~D. Cubuk, and J.~Gilmer.
\newblock A fourier perspective on model robustness in computer vision.
\newblock {\em arXiv preprint arXiv:1906.08988}, 2019.

\bibitem{WRN2016}
S.~Zagoruyko and N.~Komodakis.
\newblock Wide residual networks.
\newblock In {\em British Machine Vision Conference}, 2016.

\bibitem{zhang2017mixup}
H.~Zhang, M.~Cisse, Y.~N. Dauphin, and D.~Lopez-Paz.
\newblock mixup: Beyond empirical risk minimization.
\newblock {\em arXiv preprint arXiv:1710.09412}, 2017.

\bibitem{zhong2017random}
Z.~Zhong, L.~Zheng, G.~Kang, S.~Li, and Y.~Yang.
\newblock Random erasing data augmentation.
\newblock {\em arXiv preprint arXiv:1708.04896}, 2017.

\bibitem{zhu2017data}
X.~Zhu, Y.~Liu, Z.~Qin, and J.~Li.
\newblock Data augmentation in emotion classification using generative
  adversarial networks.
\newblock {\em arXiv preprint arXiv:1711.00648}, 2017.

\bibitem{zoph2016neural}
B.~Zoph and Q.~V. Le.
\newblock Neural architecture search with reinforcement learning.
\newblock In {\em International Conference on Learning Representations}, 2017.

\bibitem{zoph2017learning}
B.~Zoph, V.~Vasudevan, J.~Shlens, and Q.~V. Le.
\newblock Learning transferable architectures for scalable image recognition.
\newblock In {\em Proceedings of IEEE Conference on Computer Vision and Pattern
  Recognition}, 2017.

\end{thebibliography}

\clearpage
\appendix
\section{Appendix}
\begin{table*}[tbh!]
\centering
\small
\begin{tabular}{l p{12cm} p{1.35cm}}
\hline
  Operation Name & Description & Range of \\ 
                 &             & magnitudes \\ 
  \hline
  ShearX(Y) & Shear the image and the corners of the bounding boxes along the horizontal (vertical) axis with rate \emph{magnitude}.  & [-0.3,0.3]  \\ 
  TranslateX(Y) & Translate the image and the bounding boxes in the horizontal (vertical) direction by \emph{magnitude} number of pixels.   & [-150,150] \\ 
  Rotate & Rotate the image and the bounding boxes \emph{magnitude} degrees. & [-30,30] \\ 
  Equalize & Equalize the image histogram. &\\ 
  Solarize & Invert all pixels above a threshold value of \emph{magnitude}.  & [0,256] \\ 
  SolarizeAdd &  For each pixel in the image that is less than 128, add an additional amount to it decided by the magnitude. & [0,110] \\ 
  Contrast & Control the contrast of the image. A \emph{magnitude}=0 gives a gray image, whereas \emph{magnitude}=1 gives the original image.    & [0.1,1.9]\\
  Color & Adjust the color balance of the image,  in a manner similar to the controls on a colour TV set. A \emph{magnitude}=0 gives a black \& white image, whereas \emph{magnitude}=1 gives the original image.  & [0.1,1.9]\\ 
  Brightness & Adjust the brightness of the image. A \emph{magnitude}=0 gives a black image, whereas \emph{magnitude}=1 gives the original image.   & [0.1,1.9] \\
  Sharpness & Adjust the sharpness of the image. A \emph{magnitude}=0 gives a blurred image, whereas \emph{magnitude}=1 gives the original image.   & [0.1,1.9] \\ 
  Cutout~\cite{cutout2017,zhong2017random} & Set a random square patch of side-length \emph{magnitude} pixels to gray. & [0,60] \\
  BBox\_Only\_X & Apply X to each bounding box content with independent probability, and magnitude that was chosen for X above. Location and the size of the bounding box are not changed. & \\
\hline
\end{tabular}
\caption{Table of all the possible transformations that can be applied to an image. These are the transformations that are available to the controller during the search process. The range of magnitudes that the controller can predict for each of the transforms is listed in the third column. Some transformations do not have a magnitude associated with them (e.g. Equalize).}
\label{tab:ops}
\end{table*}
\begin{table*}[tbh]
\centering
\small
\begin{tabular}{lllllll}

&Operation 1 &P & M &Operation 2 &P&M\\ 
  \hline
Sub-policy 1&TranslateX&0.6&4&Equalize&0.8&10\\
Sub-policy 2&BBox\_Only\_TranslateY&0.2&2&Cutout&0.8&8\\
Sub-policy 3&ShearY&1.0&2&BBox\_Only\_TranslateY&0.6&6\\
Sub-policy 4&Rotate&0.6&10&Color&1.0&6\\
Sub-policy 5&No operation&&&No operation&&\\
\hline
\end{tabular}
\caption{The sub-policies used in our learned augmentation policy. P and M correspond to the probability and magnitude with which the operations were applied in the sub-policy. Note that for each image in each mini-batch, one of the sub-policies is picked uniformly at random. The {\it No operation} is listed when an operation has a learned probability or magnitude of 0.}
\label{tab:policy}
\end{table*}


\end{document}